\documentclass[10pt,twocolumn,letterpaper,table]{article}

\usepackage{cvpr}              % To produce the CAMERA-READY version

\definecolor{cvprblue}{rgb}{0.21,0.49,0.74}
\usepackage[pagebackref,breaklinks,colorlinks,citecolor=cvprblue]{hyperref}

\usepackage{svg}
\usepackage{graphicx}
\usepackage{amsmath}
\usepackage{amssymb}
\usepackage{booktabs}
\usepackage{wrapfig}
\usepackage{comment}
\usepackage{fontawesome5}
\usepackage{float}
\usepackage{siunitx}
\usepackage{tabularx}

\definecolor{lightfluogreen}{RGB}{187, 255, 187} 

\usepackage[pagebackref,breaklinks,colorlinks]{hyperref}

\usepackage[capitalize]{cleveref}
\crefname{section}{Sec.}{Secs.}
\Crefname{section}{Section}{Sections}
\Crefname{table}{Table}{Tables}
\crefname{table}{Tab.}{Tabs.}

\def\paperID{6584} % *** Enter the CVPR Paper ID here
\def\confName{CVPR}
\def\confYear{2024}

\begin{document}

\title{PIE-NeRF\faPizzaSlice: Physics-based Interactive Elastodynamics with NeRF}

\author{
Yutao Feng\textsuperscript{1,2}\thanks{Both authors contributed equally to this work.}\quad Yintong Shang\textsuperscript{2}\footnotemark[1]\quad Xuan Li\textsuperscript{3}\quad
Tianjia Shao\textsuperscript{1}\thanks{Corresponding author}\quad Chenfanfu Jiang\textsuperscript{3}\quad Yin Yang\textsuperscript{2}\\
\textsuperscript{1}State Key Laboratory of CAD\&CG, Zhejiang University\\
\textsuperscript{2}University of Utah\\
\textsuperscript{3}University of California, Los Angeles\\
{\tt\small fytal0n@gmail.com}\quad{\tt\small yintong.shang@utah.edu}\quad{\tt\small xuanli1@math.ucla.edu}\\
{\tt\small tjshao@zju.edu.cn}\quad{\tt\small chenfanfu.jiang@gmail.com}\quad{\tt\small yangzzzy@gmail.com}\\
\vspace{-30pt}
}

% \begin{figure*}
%   \centering 
%   \includegraphics[width = \linewidth]{fig/teaser/teaser.pdf}
%   \caption{\textbf{Pipeline overview:}~The input of PIE-NeRF is the same as other NeRF-based frameworks, which consists of a collection of images of a static scene. An adaptive Poisson disk sampling is followed to query the 3D geometry of the model, which are sparsified into $n$ Q-GMLS kernels. Integrator points are placed over the model, including centers of Q-GMLS kernels (i.e., kernel IPs). Discretization at kernels and numerical integration at IPs enable efficient synthesis of novel and physics-based elastodynamic motions. The quadratic warping scheme helps to better retrieve the color/texture of a deformed spatial position to render the final result.}
% \end{figure*}
\twocolumn[{
\renewcommand\twocolumn[1][]{#1}
\maketitle

\begin{center}
    \centering
    \captionsetup{type=figure}
    \includegraphics[width=0.95\textwidth]{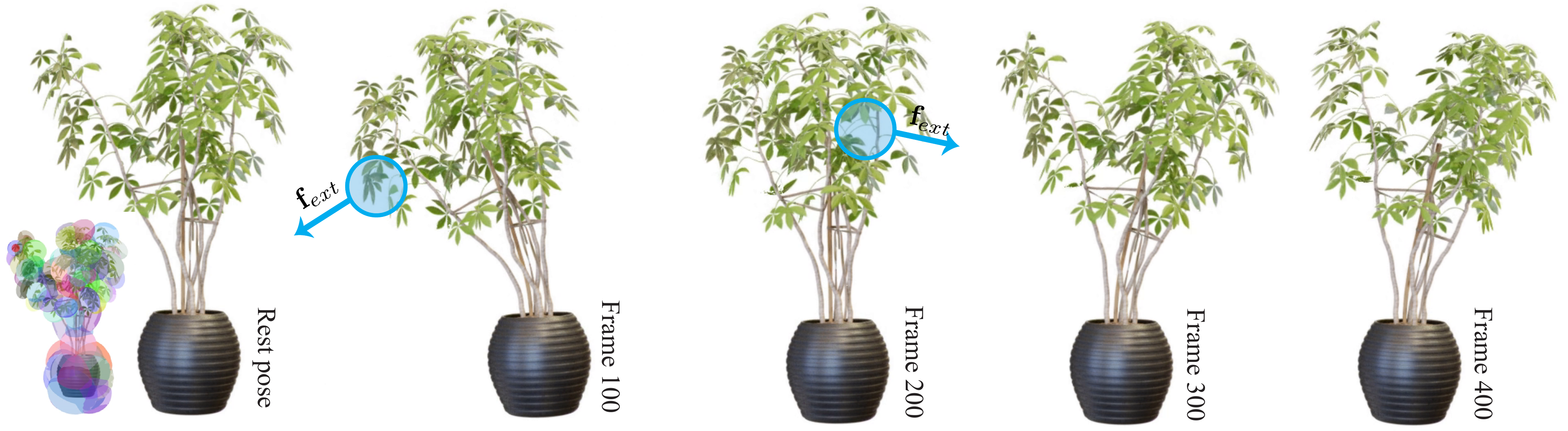}
    \captionof{figure}{\textbf{Swaying plant.}~PIE-NeRF\faPizzaSlice~~is an efficient and versatile pipeline that synthesizes physics-based novel motions of complex NeRF models interactively. In this example, the user interactively manipulates the plant by applying external forces with the mouse. The geometry of the plant is sampled in a meshless way, and a spatial model reduction is followed. We use 78 Q-GMLS kernels to capture the nonlinear dynamics of the plant in real-time. PIE-NeRF generates novel poses of the model from novel views in a physics-grounded way.}
\end{center}
}]{
  \renewcommand{\thefootnote}%
    {\fnsymbol{footnote}}
  \footnotetext[1]{Both authors contributed equally to this work}
  \footnotetext[2]{Corresponding author}
}
%%%%%%%%% ABSTRACT
\begin{abstract}We show that physics-based simulations can be seamlessly integrated with NeRF to generate high-quality elastodynamics of real-world objects. Unlike existing methods, we discretize nonlinear hyperelasticity in a meshless way, obviating the necessity for intermediate auxiliary shape proxies like a tetrahedral mesh or voxel grid. A quadratic generalized moving least square is employed to capture nonlinear dynamics and large deformation on the implicit model. Such meshless integration enables versatile simulations of complex and codimensional shapes. We adaptively place the least-square kernels according to the NeRF density field to significantly reduce the complexity of the nonlinear simulation. As a result, physically realistic animations can be conveniently synthesized using our method for a wide range of hyperelastic materials at an interactive rate. For more information, please visit \href{https://fytalon.github.io/pienerf/}{our project page}.
\end{abstract}

%%%%%%%%% BODY TEXT
\section{Introduction}\label{sec:intro}
Neural radiance field or NeRF~\cite{Mildenhall20eccv_nerf} offers a new perspective to 3D reconstruction and representation. NeRF encodes the color, texture, and geometry information of a 3D scene with an MLP net implicitly from multi-view input photos. Its superior convenience and efficacy inspired numerous follow-up research for improved visual quality~\cite{liu2020neural}, faster performance~\cite{yu2021plenoctrees,garbin2021fastnerf}, and sparser inputs~\cite{jain2021putting,yuan2022neural}. The target application has also been generalized from novel view synthesis to moving scene reconstruction or shape editing~\cite{guo2022neural,peng2022cagenerf,xu2022deforming,yuan2022nerf}. Nevertheless, complex, nonlinear, and time-coherent elastodynamic motion synthesis that is grounded on real-world physics remains less explored with the current NeRF ecosystem. 

This is probably because a physical procedure is innately incompatible with implicit representations. For dynamic models (i.e., with accelerated trajectories), spatial partial differential equations (PDEs) of stress equilibriums are coupled with an ordinary differential equation (ODE) to enforce Newtonian laws of motion. One needs a good discretization for existing simulation methods e.g., the finite element method (FEM)~\cite{zienkiewicz2005finite}, and polygonal meshes remain the most popular choice in this regard. As a result, a dedicated meshing step is often needed~\cite{yuan2022nerf,liu2023neural}. The computation cost is another concern. Dynamic simulation normally leads to a large sparse nonlinear system at each time step, and the simulation becomes expensive and has to be offline~\cite{liu2023neural}.

We propose PIE-NeRF\faPizzaSlice, a NeRF-based framework that allows users to interact with the scene in a physically meaningful way, and thus generate novel deformed poses dynamically. PIE-NeRF uses a meshless discretization scheme by adaptively sampling the density field encoded with NeRF based on the magnitude of the density gradient. The efficiency of our computation comes from the meshless spatial reduction that makes the simulation independent of the sampling resolution. 
%Conceptually, doing so is similar to using deformation handles for mesh deformation~\cite{wang2015linear} but in an implicit way. 
Specifically, PIE-NeRF employs a generalized quadratic moving least square (Q-GMLS)~\cite{martin2010unified} to drive dynamics robustly even for codimensional shapes. The prior of the quadratic displacement also allows us to design a better ray-warping algorithm so that the color/texture information of the deformed model can be accurately retrieved. PIE-NeRF uses instant neural graphics primitives (NGP)~\cite{Muller2022ngp} for faster rendering. Some noteworthy features of PIE-NeRF include:

\vspace{3 pt}
\noindent\textbf{Meshless Lagrangian dynamics in NeRF.}~We show the feasibility of integrating classic Lagrangian dynamics with NeRF in a meshless way. Honestly, we do not completely avoid the conversion from the implicit deep representation to an explicit form but a meshless shape proxy enhances the flexibility and simplifies the pipeline of simulation in NeRF. 

\vspace{3 pt}
\noindent\textbf{Robust Q-GMLS for meshless model reduction.}~We carefully design PIE-NeRF by accommodating the robustness, expressivity and computation cost simultaneously. With spatial reduction based on the Voronoi partition, we enhance the expressivity of our reduced model using quadratic displacement interpolation, which captures the nonlinear deformation of the model without locking artifacts. The quadratic field also contributes an improved ray-warping algorithm during the view synthesis.

\vspace{3 pt}
\noindent\textbf{Versatile simulation at an interactive rate.}~Being a full physics-based pipeline, PIE-NeRF can faithfully designate material parameters such as Young's modulus and Poisson's ratio to NeRF models. It is also efficient, allowing an interactive interaction between the virtual NeRF scene and end users. Thanks to our high-order interpolation, codimensional models can also be well handled with PIE-NeRF.

\section{Related work}
\label{sec:related}
\noindent\textbf{NeRF editing.}~
%NeRF~\cite{Mildenhall20eccv_nerf} offers various advantages of capturing complex 3D geometries and lighting conditions in a continuous, compact, and viewpoint-independent manner, thereby enabling high-quality novel view synthesis. 
Since the advent of NeRF, many techniques tailored for implicit representation have emerged i.e.,
%For dynamic scene generation, we focus on the cases where non-rigid deformation is dominant and thus shape deformation should be estimated. 
discretely estimating the deformation/displacement field for each frame ~\cite{park2021nerfies,park2021hypernerf,tretschk2021non,weng2022humannerf} or estimating the time-continuous 3D motion field~\cite{pumarola2021d,du2021neural,li2021neural,xian2021space,gao2021dynamic,liu2022devrf,guo2023forward}. Recently, Cao and colleagues have matched space-time features to hexplanes for improved NeRF training~\cite{cao2023hexplane}. 

There also exists a wide range of NeRF editing methods for various purposes. These include semantic-driven editing~\cite{wang2022clip,bao2023sine,song2023blending,dong2023vica,haque2023instruct,mikaeili2023sked}, shading-driven adjustments (like relighting and texturing)~\cite{srinivasan2021nerv,liu2021editing,rudnev2022nerf,wu2022palettenerf,ye2023intrinsicnerf,gong2023recolornerf}, scene modifications (such as object addition or removal)~\cite{zhang2021editable,yang2021learning,kobayashi2022decomposing,weder2023removing,lazova2023control}, face editing~\cite{sun2022fenerf,zhang2022fdnerf,jiang2022nerffaceediting,hwang2023faceclipnerf}, physics based editing from video\cite{qiao2022neuphysics, hofherr2023neural}, and multi-purpose editing~\cite{yang2022neumesh,wang2023seal,jambon2023nerfshop}.

Geometry editing with NeRF has also been widely investigated~\cite{yuan2022neural,kania2022conerf,zheng2023editablenerf,yuan2023interactive}. They normally concern static shapes only, where the as-rigid-as-possible (ARAP) energy suffices~\cite{sorkine2007rigid} in most situations. The ARAP energy is often computed by converting neural implicit representation to some explicit forms like grid or mesh~\cite{garbin2022voltemorph}. To reduce the computational overhead, some opt for coarser meshes, utilizing cage-based deformation techniques~\cite{peng2022cagenerf,xu2022deforming,jambon2023nerfshop}.

% such as a volumetric tetrahedral mesh~\cite{hu2018tetrahedral}.
% Instead of directly deforming the implicit representation as we propose, a less ideal method involves converting the implicit geometry to an explicit mesh, on which many traditional deforming methods can be employed. This is typically done using meshing algorithms like Tetwild~\cite{hu2018tetrahedral}. While it is possible to generate a fine-resolution tetrahedral mesh suitable for generic, real-time, and controllable deformations~\cite{garbin2022voltemorph}, the more common approach opts for coarser meshes, utilizing cage-based deformation techniques~\cite{peng2022cagenerf,xu2022deforming,jambon2023nerfshop}.

Point-based shape editing is a viable alternative. Chen and colleagues~\cite{chen2023neuraleditor} proposed a more general editing framework with the optimized points inherent in a point-based variant of NeRF~\cite{xu2022point}. More recently, Prokudin and colleagues exploited a point-based surface derived from an implicit volumetric representation~\cite{prokudin2023dynamic}. 
% In a similar vein, our method is also capable of incorporating frequently-used hyperelastic models such as as-rigid-as-possible (ARAP)~\cite{sorkine2007rigid} and Neo-Hookean~\cite{smith2018stable}.

\vspace{3 pt}
%-------------------------------------------------------------------------
\noindent\textbf{Physics-based deformable model.}
The concept of deformable model dates back to 1980s~\cite{terzopoulos1987elastically}, primarily associated with physics-based models~\cite{nealen2006physically}. Typically, an explicit discretization is needed such as mass-spring systems~\cite{baraff1998large} that were widely used in early graphics applications. FEM has become the standard for physics-based simulation~\cite{bro1996real,muller2004interactive,sifakis2012fem,kim2022dynamic}, wherein the deformation is usually measured by integrating over each tetrahedral or hexahedral unit, that is, the element~\cite{zienkiewicz2005finite}. Physics-based modeling is known to be expensive, which inspires a series of research for accelerated simulation such as model reduction~\cite{barbivc2005real} or GPU parallelization~\cite{wang2016descent}.
%In the realm of shape editing methods, both mesh-based and meshless methods have been developed. For a comprehensive review, see ~\cite{yuan2021revisit}.

\vspace{3 pt}
%-------------------------------------------------------------------------
\noindent\textbf{Meshless simulation.}
Meshless methods use unstructured vertices in lieu of a predefined mesh~\cite{belytschko1996meshless,liu2003mesh,fries2003classification}. This modality is quite effective when the simulation domain has varying topology, such as fluids, gases, fracturing or melting. Meshless deformation has evolved to handle continuum mechanics~\cite{muller2004point}. A notable example is the shape matching~\cite{muller2005meshless}. With the core idea of constraining vertices' positions, shape matching paves the way to the position-based methods~\cite{muller2007position,steinemann2008fast,muller2011solid}.  Similar to shape functions in FEM, meshless methods also need well-designed interpolation schemes~\cite{fries2003classification}, such as moving least squares (MLS)~\cite{pauly2003shape,muller2004point} or smoothed-particle hydrodynamics SPH~\cite{antoci2007numerical}. 
% The inherent flexibility of meshless methods facilitates versatile simulations across a variety of object types. Macklin et al.~\cite{macklin2014unified} build a system based on Position-Based Dynamics (PBD) where all objects are represented as particles to achieve unified simulation. Martin et al.~\cite{martin2010unified} encode linearized strain in a local proxy to simulate three, two and one-dimensional solids in a unified way, demonstrating the adaptability of meshless methods, which our work will build upon.
%

Due to the large volume of relevant work, we can only discuss a small fraction of excellent prior arts in this section. Nevertheless, we note that synthesizing novel dynamic motions of a NeRF scene in a physically grounded way remains less explored. This gap inspires us to develop PIE-NeRF\faPizzaSlice. PIE-NeRF is a physics-based, meshless, and efficient framework allowing users to interactively manipulate the NeRF scene.
\begin{figure*}
  \centering 
  \includegraphics[width = \linewidth]{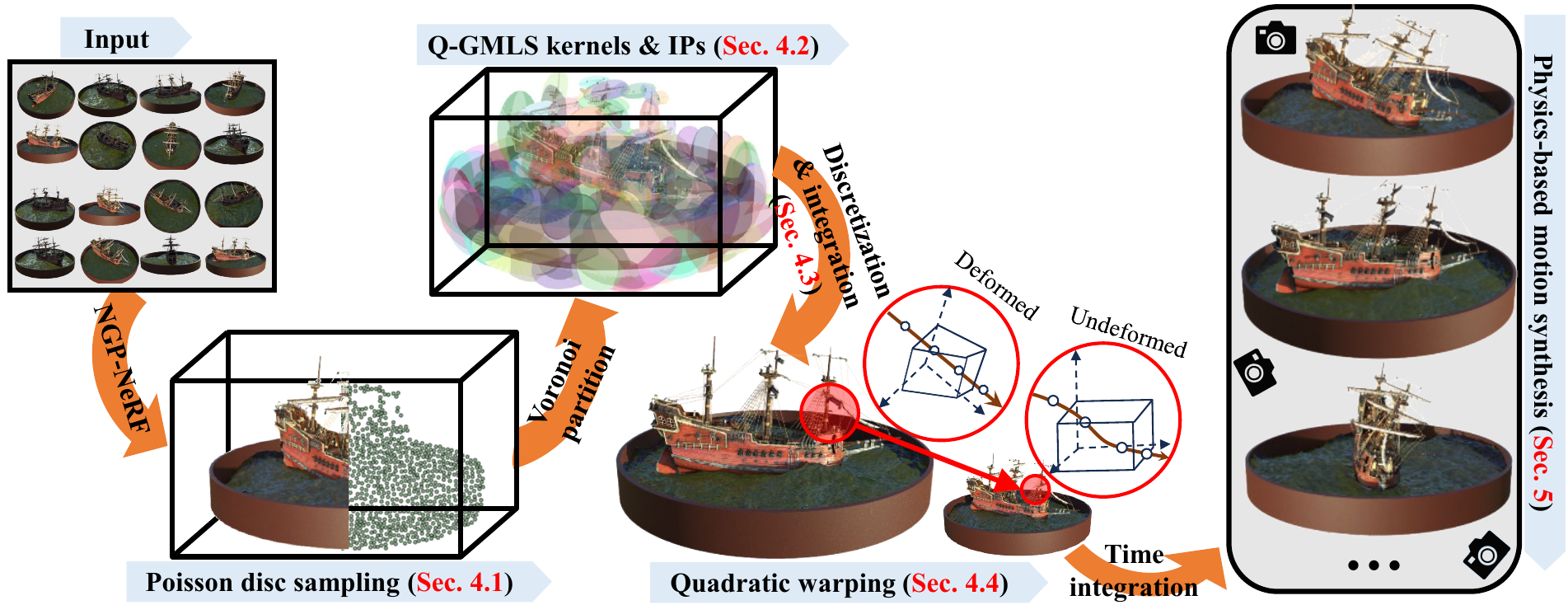}
  \caption{\textbf{Pipeline overview.}~The input of PIE-NeRF is the same as other NeRF-based frameworks, which consists of a collection of images of a static scene. An adaptive Poisson disk sampling is followed to query the 3D geometry of the model, which are sparsified into $n$ Q-GMLS kernels. Integrator points are placed over the model, including centers of Q-GMLS kernels (i.e., kernel IPs). Discretization at kernels and numerical integration at IPs enable efficient synthesis of novel and physics-based elastodynamic motions. The quadratic warping scheme helps to better retrieve the color/texture of a deformed spatial position to render the final result.}
  \label{fig:overview}
\end{figure*}

\section{Preliminary}
\label{sec:pre}
To make the paper self-contained, we start with a brief review of some core techniques on which our pipeline is built. More details of our system are elaborated in \S~\ref{sec:method}. 
\subsection{Neural radiance field}
NeRF implicitly represents the geometry and appearance information of a 3D scene via a multi-layer perceptron (MLP) net. Given the camera parameters, a pixel's color on the image plane is obtained via integrating the density and color along the ray.
% \begin{equation} \label{eq:nerf}
% \widehat C(\mathbf r) = \sum_{i} \exp (-\sum_{j} \sigma_j\delta_j) (1 - \exp (-sigma_i\delta_i)) \mathbf c_i.
% \end{equation}
A spatial coordinate $\mathbf{p}$ and a ray direction $\mathbf{d}$ are often encoded as a feature vector $\psi_{\mathbf p}, \psi_{\mathbf d}$ before being fed to the MLP for the prediction of density ($\sigma$) and color ($\mathbf c$).
%such as: $F_{\Theta}: (\psi_{\mathbf p}, \psi_{\mathbf d})\rightarrow (\sigma, \mathbf c)$. 
For instance, the vanilla NeRF~\cite{Mildenhall20eccv_nerf} uses positional encoding to better tackle high-frequency information with MLPs. Our pipeline uses the instant neural graphics primitives (NGP)~\cite{Muller2022ngp}. NGP adopts a multi-level hash-based encoding scheme and has demonstrated a strong performance in terms of both efficiency and quality. 
% With NGP, the color prediction of a ray $\mathbf r(t)$ becomes:

% where $\delta_i=t_{i+1}-t_{i}$ is the sampling interval. 
%The network is optimized by minimizing the $\mathcal{L}^2$ loss between the ground truth color $C(\mathbf r)$ and the predicted color $\hat C(\mathbf r)$. 

% Some known encoding schemes include  positional encoding~\cite{mildenhall2021nerf}, 
% instant neural graphics primitives\cite{Muller2022ngp}
% For instance, the vanilla NeRF~\cite{mildenhall2021nerf} uses positional encoding
% There are various options options for the embedded function $\psi$, including traditional NeRF \cite{mildenhall2021nerf}, which uses positional encoding, and instant neural graphics primitives\cite{Muller2022ngp}, which employ multiresolution hash encoding. The ray $\mathbf r(t)$ 's color $\hat C(\mathbf r)$ then can be calculated as:

% \begin{equation}
% \hat C(\mathbf r) = \sum_{i=1}^N \mathrm{exp(-\sum_{j=1}^{i-1}\sigma_j\delta_j)}(1-\mathrm{exp(-\sigma_i\delta_i)})\mathbf c_i,
% \end{equation}
% where $\delta_i=t_{i+1}-t_{i}$ is the distance between adjacent samples. The network is optimized by minimizing the $\mathcal{L}^2$ loss between the ground truth color $C(\mathbf r)$ and the predict color $\hat C(\mathbf r)$. 

\subsection{Nonlinear elastodynamic}
Following the classic Lagrangian mechanics~\cite{murray1997nonlinear}, the dynamic equilibrium of a 3D model is characterized as: 
\begin{equation}\label{eq:lagrangian}
\frac{\mathrm{d} }{\mathrm{d} t}
\left(\frac{\partial L}{\partial \dot{\mathbf{q}}}\right) - \frac{\partial L}{\partial \mathbf{q}} = \mathbf{f}_q,
\end{equation}
where $L = T - U$ is \emph{Lagrangian} i.e., the difference between the kinematic energy ($T$) and the potential energy ($U$) of the system. $\mathbf{q}$ and $\dot{\mathbf{q}}$ are generalized coordinate and velocity. $\mathbf{f}_q$ is the generalized external force. Given a time integration scheme such as implicit Euler: $\mathbf{q}_{n+1} = \mathbf{q}_n + h\dot{\mathbf{q}}_{n+1}$, $\dot{\mathbf{q}}_{n+1} = \dot{\mathbf{q}}_n + h\ddot{\mathbf{q}}_{n+1}$, Eq.~\eqref{eq:lagrangian} can be reformulated a set of nonlinear equations to be solved at each time step:
\begin{equation}\label{eq:motion}
    \mathbf{M}(\mathbf{q}_{n+1} - \mathbf{q}_n - h \dot{\mathbf{q}}) = h^2 \left(-\frac{\partial U}{\partial \mathbf{q}} + \mathbf{f}_q\right).
\end{equation}
Here, the subscript indicates the time step index, and $h$ is the time step size. $\mathbf{q}_{n+1}$ is the unknown system coordinate to be solved, while all the kinematic variables of the previous time step such as $\mathbf{q}_n$ or $\dot{\mathbf{q}}_n$ are considered known. $-\partial U/ \partial \mathbf{q}$ is the negative gradient of the potential, which embodies the internal force. %As to be detailed in \S~\ref{sec:method}, we use Q-GMLS to assemble Eq.~\eqref{eq:motion} without meshing or voxelizing the NeRF field. 

 \section{Our method}\label{sec:method}
As shown in Fig.~\ref{fig:overview}, the input of our system is a collection of images of a given 3D scene. We use NGP to encode positional and texture information and train the corresponding NeRF. Afterwards, we disperse particles into the scene. Those particles form an unstructured point-cloud-like proxy of the 3D model of interest. They are then grouped under a Voronoi partition, and the centers of Voronoi cells house the generalized coordinate of the system ($\mathbf{q}$ in Eq.~\eqref{eq:motion}). We further assign multiple integrator points (IPs) to facilitate energy integration. A quadratic generalized moving least square (Q-GMLS) strategy is used to discretize the Lagrangian equation of Eq.~\eqref{eq:lagrangian}. 
%The computational complexity of our system depends on the total number of Q-GMLS kernels, and we find that a few kernels are often sufficient to well describe nonlinear motions of complex shapes. 
With the help of GPU, the simulation can be done at an interactive rate or even in real-time. We leverage the deformation information at IPs to infer the rest-pose position during NGP-based NeRF rendering. Thanks to NGP, this procedure is also in real-time. Our pipeline allows users to interact with a NeRF scene by applying external forces, position constraints etc., leading to novel and physics-grounded dynamic effects. Next, we give detailed expositions of each major step of the pipeline.

% We present a complete meshless pipeline of simulating and rendering 3D models without the use of traditional meshes or tetrahedral meshes. Our model enables user to interact with the 3D model and generates elastic dynamics from different views. We will start with the \textit{Neural Radiance Field} (NeRF)\cite{mildenhall2021nerf}, which is able to reconstruct the 3D model's geometry and appearance from a set of multi-view images. We then present how to exact the particles for our meshless simulation according to the scene's density field. After obtaining the particle representation of the 3D model, we introduce the \textit{Generalized Moving Least Squares} (GMLS) \cite{martin2010unified} technique to efficiently and accurately simulate elastic rods, shells and solids. Finally, we propose a quadratic estimation method to map points in the deformed scene back to their undeformed positions, allowing us to query their density and color information in the NeRF. 

\begin{figure}
  \centering 
  \includegraphics[width=1\linewidth]{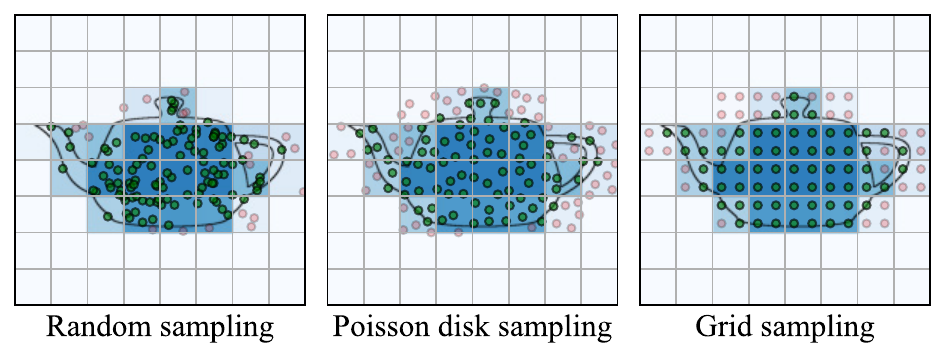}
  \caption{\textbf{Particle sampling.}~Our method is compatible with most sampling algorithms -- as long as particles cover the shape of the 3D model sufficiently well. In our implementation, we design a novel augmented Poisson disk sampling scheme that is fast and well captures the boundary of the model by default.}
  \label{fig:sampling}
  \vspace{-3pt}
\end{figure}

\subsection{Augmented Poisson disk sampling}\label{subsec:sample}
After the NGP-NeRF is trained, we choose a meshless way to model the geometry of the 3D shape. While the underlying goal of this step is similar to other static NeRF editing systems~\cite{yuan2023interactive,xu2022deforming,peng2022cagenerf}, being mesh-free makes our pipeline more flexible and versatile. In theory, any sampling method should work as long as the sampling particles sufficiently capture the boundary of the model. For instance, one can distribute particles by simply following the evenly-spaced grid (Fig.~\ref{fig:sampling}, right). Doing so is similar to using a grid-based cage to approximate the shape of the model~\cite{guo2022neural}.

Alternatively, we design an augmented Poisson disk sampling (PDS) strategy. The original PDS requires that the distance between any two particles be larger than a threshold $\bar{r}$. Starting from an initial point, PDS then tries to fill a banded ring between $\bar{r}$ and $2\bar{r}$ with new samples i.e., see~\cite{Bridson2007PoissonSampling}. Our observation is that more particles are needed at the boundary of the shape, which coincides with a sharp density variation. To this end, we adaptively adjust the sample radius $r$ based on the norm of the density gradient of NGP-NeRF $\| \nabla \sigma \|$ such that:
\begin{equation}\label{eq:raidus}
r = \min \left\{\bar{r}, \kappa \frac{\bar{r}}{\sqrt{\| \nabla \sigma \| + \alpha}} \right\},
\end{equation}
where $\alpha = 10^{-3}$ is a small number avoiding the division-by-zero error. Eq.~\eqref{eq:raidus} suggests that the actual sample radius $r$ decreases when $\| \nabla \sigma \|$ is a large quantity. The density gradient $\| \nabla \sigma \|$ can be conveniently computed by differentiating the NPG-NeRF using \texttt{AutoDiff}~\cite{paszke2017automatic}. We discard PDS particles whose density values are less than $\epsilon = 10^{-2}$, which are visualized as pink dots in Fig.~\ref{fig:sampling}. 

Our sampling ensures that the distance between a PDS particle at $\mathbf{x}$ and its nearest neighbor is at least $r(\mathbf{x})$, and we assign a volume of the PDS particle as:
\begin{equation}
V(\mathbf{x}) = \frac{4}{3}\pi r^3(\mathbf{x}).    
\end{equation}

% we need to represent the 3D model we intend to simulate using a set of particles. Initially, We sample points in the entire scene using arbitrary sample strategy and then remove particles with low values of density which contribute minimally to the volume rendering. Specifically, we set $\epsilon=10^{-2}$ as a constant threshold that we will remove particles with lower density value than $\epsilon$. In our work, we use \textit{Poisson Disk Sampling} \cite{Bridson2007PoissonSampling} which can evenly sample particles in the scene. Moreover, we adaptively set the Poisson Disk Sampling radius based on the gradient of density to extract a more accurate boundary of the object. Fig.~\ref{fig:PoissonDiskSampling} compares our adaptive method with random sampling and grid sampling.

\subsection{Q-GMLS kernels and integrator points}
\begin{wrapfigure}[6]{r}{0.4\linewidth}
    %\centering 
    \vspace{-14 pt}
    \hspace{-20 pt}
    \includegraphics[width=\linewidth]{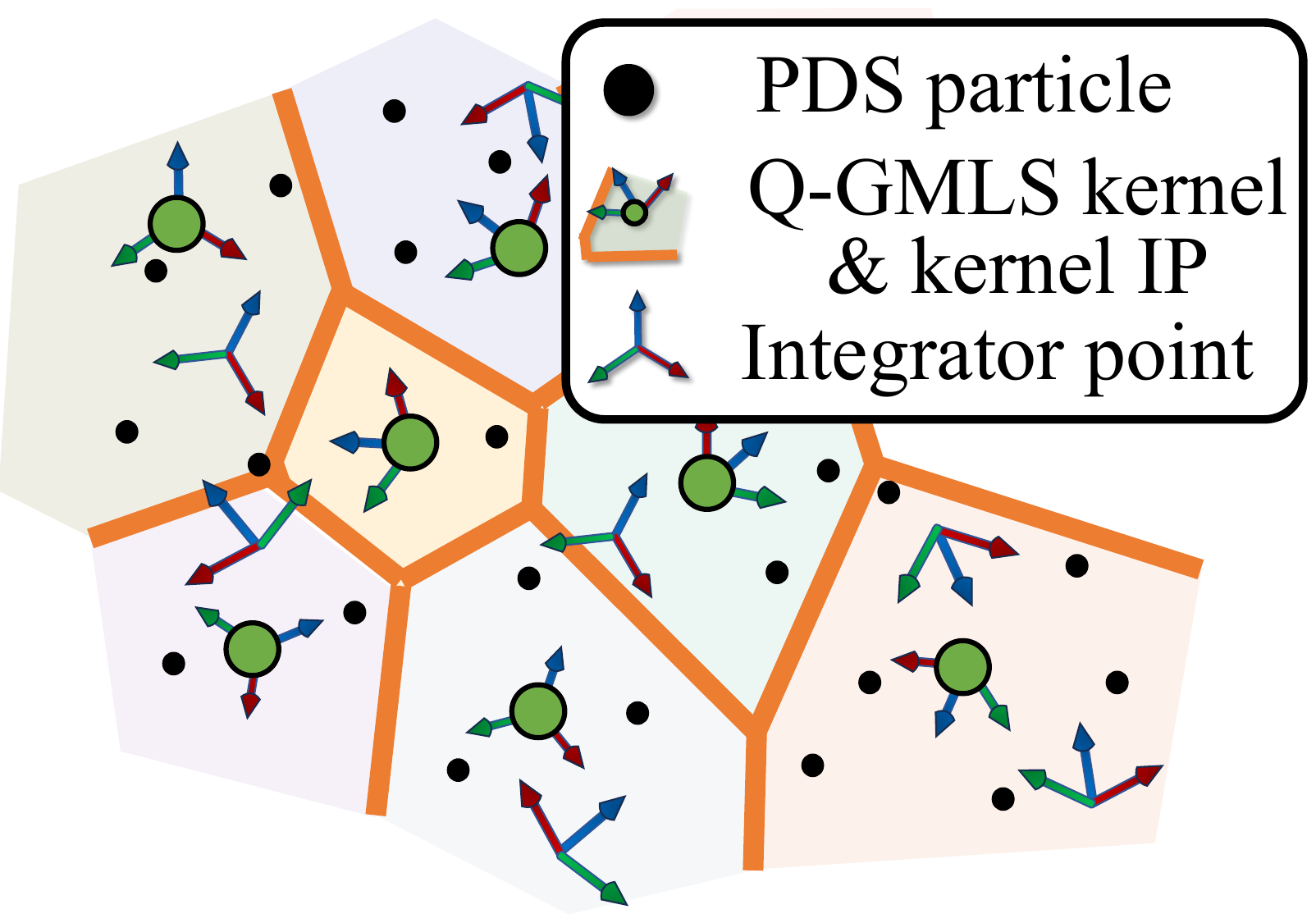}
    %\caption{\textbf{Voronoi clustering:}~A Voronoi tessellation offers good neighborhood information of PDS particles (black points). In addition to the centers of Voronoi cells (green dots),  additional IPs (small frames) are deployed to facilitate the Q-GMLS integration.}\label{fig:voronoi}
\end{wrapfigure}
We perform a Voronoi partition~\cite{aurenhammer1991voronoi} over PDS particles (see the inset) and use each Voronoi cell as a GMLS kernel for the body-wise displacement interpolation to reduce the computation overhead. 
%The integration is, on the other hand, performed at IPs, which serves as 
% Let $\Omega$ be the body of a 3D model in NeRF sampled by PDS particles, whose rest-shape positions are $\mathbf{x}_i\in\mathbb{R}^3, i = 1,2,...,n$. They are displaced by $\mathbf{u}_i $ under dynamical deformations. It is natural to assume the displacement field is  w.r.t $\mathbf{x}_i$ such that $\mathbf{u}_i = $
Let $\Omega$ be the body of a 3D model in NeRF sampled by PDS particles with $n$ GMLS kernels.
%and $\mathbf{u}(\mathbf{x})$ be the displacement field of rest position $\mathbf{x}$ on $\Omega$. 
%We use $n$ GMLS kernels to discretize the body-wise displacement field (instead of directly using PDS particles). $\mathbf{u}_i$ and $\mathbf{x}_i$ denote the displacement and rest position of the $i$-th kernel's center (green dots in Fig.~\ref{fig:voronoi}), and 
The classic MLS assumes a kernel possesses an affine displacement filed: $\mathbf{A}_i\mathbf{p}(\mathbf{x}_i)$, where $\mathbf{x}_i$ is the center of the $i$-th kernel at the rest pose (green dots in the inset); $\mathbf{p}(\mathbf{x})=[1,\mathbf{x}^\top]^\top$. 
%The displacement at an arbitrary material point $\mathbf{x} \in \Omega$ is considered as a weighted average of all kernels, and 
By minimizing a displacement-based target of:
$\sum_{i=1}^n w(\mathbf{x} -\mathbf{x}_i)\left\|\mathbf{A}\mathbf{p}(\mathbf{x}_i) - \mathbf{u}_i\right\|^2$ one can obtain:
\begin{equation}\label{eq:affine}
\mathbf{u}(\mathbf{x})=\sum_{i=1}^n\mathbf{u}_iN_i(\mathbf{x}).
\end{equation}
Here $\mathbf{u}_i = \mathbf{u}(\mathbf{x}_i)$ is the displacement of $i$-the kernel center;
%We can then obtain the body-wise displacement interpolation in the standard form of:
% Let $\mathbf{a} = \text{vec}(\mathbf{A}^\top)\in\mathbb{R}^{12}$ be the vectorized coefficients of this affine function. 
% The optimal coefficients of $\mathbf{a}$ i.e., minimizing $\sum_{i=0}^n w(\mathbf{x} -\mathbf{x}_i)\left\|\mathbf{A}\mathbf{p}(\mathbf{x}_i) - \mathbf{u}_i\right\|^2$ can be analytically computed, where $\mathbf{u}_i$ and $\mathbf{x}_i$ denote the displacement and rest position of the $i$-th kernel's center (green dots in Fig.~\ref{fig:voronoi}).
$N_i(\mathbf{x})=\mathbf{p}(\mathbf{x})^\top\mathbf{G}^{-1}(\mathbf{x})\mathbf{p}(\mathbf{x}_i)w(\mathbf{x}-\mathbf{x_i})$, for $\mathbf{G}(\mathbf{x})=\sum_{i=1}^n w(\mathbf{x}-\mathbf{x}_i)\mathbf{p}(\mathbf{x}_i)\mathbf{p}(\mathbf{x}_i)^\top$, is a shape-function-like trial function; and $w(\mathbf{d})=(1-\left\| \mathbf{d} \right\|^2)^3$ is a MLS weighting function based on the distance between $\mathbf{x}$ and $\mathbf{x}_i$.

%Both can be pre-computed.

% For any point $\mathbf{x}_0\in\Omega$, we want the displacement field $\mathbf{u}(\mathbf{x})$ in its vicinity can be represent as a polynomial $\mathbf{a}^T\mathbf{p}(\mathbf{x})$ with a vector of monomials $\mathbf{p}(x,y,z)=(1,x,y,z)^T$ and the coefficients $\mathbf{a} = \mathbf{a}(\mathbf{x}_0)$. Our goal lead to a local least square fitting to the displacement $u_i$ by minimizing

% \begin{equation}\label{eq:MLS}
% J(\mathbf{a})=\sum_{i=0}^n w(\mathbf{x}_0-\mathbf{x}_i)\left\|\mathbf{a}^T\mathbf{p}(\mathbf{x}_i) - \mathbf{u}_i\right\|^2
% \end{equation}

% where $w(\mathbf{x}_0-\mathbf{x}_i)$ is the weighting kernel that we apply $w(\mathbf{d})=(1-\left\|d\right\|^2)^3$ in our implementation. Anylutically minimizing $J(\mathbf{a})$ by setting $\frac{\partial J}{\partial \mathbf{a}}=\mathbf{0}$ yields
As the complexity of the simulation is up to $n$, we are in favor of using fewer kernels for faster computation. Doing so is likely to have $\mathbf{x}_i$ be colinear/coplanar, and $\mathbf{G}$ becomes singular. To improve the robustness of the kinematic interpolation of Eq.~\eqref{eq:affine}, GMLS takes the local deformation gradient information into account, which seeks the optimal $\mathbf{A}_i$ to minimize $\sum_{i=1}^n w(\mathbf{x} -\mathbf{x}_i)\left\|\mathbf{A}_i\mathbf{p}(\mathbf{x}_i) - \mathbf{u}_i\right\|^2 + \sum_{i=1}^n \sum_{j=1}^3 w(\mathbf{x}_0-\mathbf{x}_i)\left\|\mathbf{A}_i\mathbf{p}_{,j}(\mathbf{x}_i) - \mathbf{u}_{i,j}\right\|^2$. The comma here denotes the partial differentiation such that $\mathbf{u}_{i, 1} = \partial \mathbf{u}_i / \partial x$, $\mathbf{u}_{i, 2} = \partial \mathbf{u}_i / \partial y$, and $\mathbf{u}_{i, 3} = \partial \mathbf{u}_i / \partial z$.

For thin and codimensional shapes, affine GMLS suffers from locking issues, wherein linearized shearing energy becomes orders stronger than nonlinear bending/twisting due to the interpolation error. This problem gets more serious with fewer kernels. To this end, we elevate the interpolation order, leading to quadratic GMLS or Q-GMLS, which assumes the per-kernel displacement field is quadratic. Namely, each $x$, $y$, or $z$ component of the displacement (i.e., for $j = 1,2,3$ respectively) is fit by: $u_{ij} = \mathbf{x}_i^\top \mathbf{Q}^j_i \mathbf{x}_i + \mathbf{a}_i^{j\top} \mathbf{p}(\mathbf{x}_i)$. Here $\mathbf{Q}_i^j$ is a symmetric tensor, and $\mathbf{a}_i^j\in\mathbb{R}^4$ is $j$-th row of $\mathbf{A}_i$. The Q-GMLS displacement interpolation can then be derived as:
% However, the above equation of classic MLS needs the matrix $\mathbf{G}(\mathbf{x})$ to be invertible, which means there have to be sufficient kernel $\mathbf{x}_i$ surpoorting $\mathbf{x}$ and $\mathbf{x}_i$ must not be coplanar. GMLS solved this problem by adding $J(\mathbf{a})$ in \eqref{eq:MLS} the high order derivative error term, e.g. first order $\sum_{i=0}^n \sum_{j=1}^3 w(\mathbf{x}_0-\mathbf{x}_i)\left\|\mathbf{a}^T\mathbf{p}_{,j}(\mathbf{x}_i) - \mathbf{u}_{i,j}\right\|^2$, where comma denotes partial differentiation, e.g., $\mathbf{u}_{,i}\equiv \partial \mathbf{u} / \partial \mathbf{e}_i$. 
% In our implementation, we use the quadratic GMLS(Q-GMLS) and \eqref{eq:MLSopt} turns to
\begin{multline}\label{eq:qgmls}
\mathbf{u}(\mathbf{x})=\sum_{i=1}^n\big[\mathbf{u}_iN_i
+\sum_{j}\mathbf{u}_{i,j}N_i^j 
+\sum_{j,k}\mathbf{u}_{i,jk}N_i^{jk}\big],
\end{multline}
for $j,k = 1,2,3$. Here
\begin{equation}\label{eq:N}
\begin{split}
N_i(\mathbf{x})&=\mathbf{p}^\top(\mathbf{x})\mathbf{G}^{-1}(\mathbf{x})\mathbf{p}(\mathbf{x}_i)w(\mathbf{x}-\mathbf{x}_i),\\
N_i^j(\mathbf{x})&=\mathbf{p}^\top(\mathbf{x})\mathbf{G}^{-1}(\mathbf{x})\mathbf{p}_{,j}(\mathbf{x}_i)w(\mathbf{x}-\mathbf{x}_i),\\
N_i^{jk}(\mathbf{x})&=\mathbf{p}^\top(\mathbf{x})\mathbf{G}^{-1}(\mathbf{x})\mathbf{p}_{,jk}(\mathbf{x}_i)w(\mathbf{x}-\mathbf{x}_i)
\end{split}
\end{equation}
only depend on the rest-shape position $\mathbf{x}$, and
\begin{multline}
\mathbf{G}(\mathbf{x})=\sum_{i=1}^n w(\mathbf{x}-\mathbf{x}_i) \left[\mathbf{p}(\mathbf{x}_i)\mathbf{p}^\top(\mathbf{x}_i)\right.\\ 
+\sum_{j}\mathbf{p}_{,j}(\mathbf{x}_i)\mathbf{p}_{,j}^\top(\mathbf{x}_i) +\sum_{j,k} \left.\mathbf{p}_{,jk}(\mathbf{x}_i)\mathbf{p}^\top_{,jk}(\mathbf{x}_i)\right].   
\end{multline}
It is convenient to re-organize Eq.~\eqref{eq:qgmls} as:
\begin{equation}\label{eq:u}
    \mathbf{u}(\mathbf{x}) = \mathbf{J}(\mathbf{x})\mathbf{q},
\end{equation}
such that $\mathbf{J} = [N_1 \mathbf{I}, N_1^1 \mathbf{I}, N_1^2 \mathbf{I},...,N_1^{11} \mathbf{I},...] \in\mathbb{R}^{3 \times 30n}$ and $\mathbf{q} = [\mathbf{u}_1^\top, \mathbf{u}_{1,1}^\top,\mathbf{u}_{1,2}^\top,...,\mathbf{u}_{1,11}^\top,...]^\top \in\mathbb{R}^{30n}$ are the Jacobi matrix and generalized coordinate (i.e., in Eq.~\eqref{eq:lagrangian}). Thus the generalized external force is computed via: $\mathbf{f}_q = \mathbf{J}^\top\mathbf{f}_{ext}$. 

\subsection{Energy integration}
The total kinematic and potential energies of the model are:
\begin{equation}\label{eq:int}
T = \frac{1}{2}\int_\Omega \rho(\mathbf{x}) \dot{\mathbf{x}}^\top\dot{\mathbf{x}} \mathrm{d}\Omega, \;\text{and} \;\; U = \int_\Omega \Psi(\mathbf{x}) \mathrm{d}\Omega.
\end{equation}
%While Eq.~\eqref{eq:u} allows us to use generalized coordinate $\mathbf{q}$ to prescribe the kinematic configuration of the model, the energy integrals are also needed to assemble the system. 
We want to avoid integrating over all the PDS particles. Therefore, our system includes another set of integrator points or IPs. Conceptually, IPs are similar to the quadrature points used in numerical integration~\cite{gerstner1998numerical}, which allows us to substantially reduce the computational cost of full integrals in Eq.~\eqref{eq:int}. In addition to the centers of Q-GMLS kernels i.e., kernel IPs, we add more IPs aiming to approximate Eq.~\eqref{eq:int} with high accuracy. Specifically, we initialize new IPs at the PDS particle which is the most distant from existing IPs to sample remote $T$ and $V$ values. This strategy however tends to favor PDS particles at the model's boundary. As a result, we apply a few Lloyd relaxations~\cite{du2006convergence} to new IPs while keeping kernel IPs fixed. The total number of IPs is bigger than the number of Q-GMLS kernels but they are of the same order, and we use $\mathcal{I}$ to denote the set of all the IPs.

% Let $m$ be the number of IPs. 
% When numerically integrating $T$ and $V$,
\subsection{Per-IP integration}
\begin{wrapfigure}[5]{r}{0.27\linewidth}
\vspace{-14 pt}
\hspace{-20 pt}
    %\centering 
    \includegraphics[width=\linewidth]{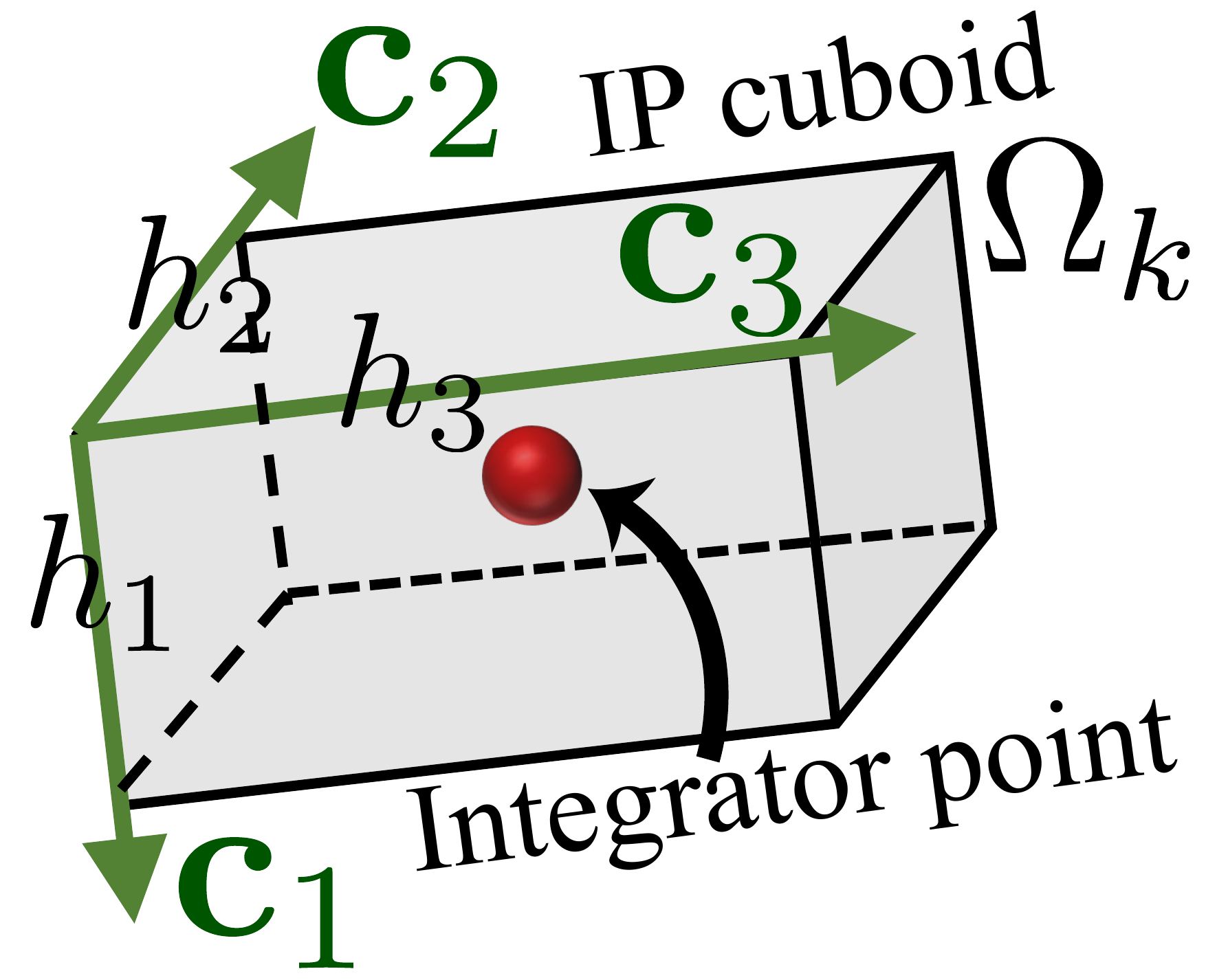}
    % \caption{\textbf{Integrator point:}~IPs are small cuboids accumulating mass and energies of nearby PDS particles.}\label{fig:ip}
\end{wrapfigure}
We envision each IP as a small elastic cuboid $\Omega_k$ (see the inset) with three edges being $\mathbf{c}_1$, $\mathbf{c}_2$, $\mathbf{c}_3$ whose lengths are $h_1$, $h_2$, $h_3$ respectively. Its covariance matrix can be computed as: $\mathbf{C} = \sum_j V(\mathbf{x}_j)\mathbf{x}_j \mathbf{x}_j^\top$, where the summation carries over K-nearest PDS particles whose rest positions are $\mathbf{x}_j$. Being a symmetric matrix, $\mathbf{C}$ always has three real non-negative eigen values namely, $\lambda_1$, $\lambda_2$, and $\lambda_3$. Note that coplanar geometry around an IP can make $\mathbf{C}$ singular. It is fine for numerical integration, suggesting the strains along certain directions are zero. We then set the ratio among $h_i$ to be the same as $\sqrt{\lambda_i}$ (i.e., $h_1 : h_2 : h_3$ equals $\sqrt{\lambda_1} : \sqrt{\lambda_2} : \sqrt{\lambda_3}$)  while requiring $\Pi_i h_i = \sum_j V(\mathbf{x}_j)$. Those two constraints allow us to compute $h_1$, $h_2$, and $h_3$ while $\mathbf{c}_i$ are the corresponding eigen vectors. 

% They can be computed using the covariance matrix of its Voronoi region (as the result of Lloyd iterations). ($\lambda_i$ is the eigen values of the covariance matrix, $h_1:h_2:h_3=\sqrt{\lambda_1}:\sqrt{\lambda_2}:\sqrt{\lambda_3}$, $h_1h_2h_3$ is equal to the sum of volume of particles in the Voronoi region, $\mathbf{c}_1$,$\mathbf{c}_2$,$\mathbf{c}_3$ are the corresponding eigen vectors)

The total kinematic energy can now be approximated as:
\begin{multline}
T = \frac{1}{2}\int_\Omega \rho(\mathbf{x}) \dot{\mathbf{x}}^\top\dot{\mathbf{x}} \mathrm{d}\Omega = \frac{1}{2} \dot{\mathbf{q}}^\top\left(\int_\Omega \rho \mathbf{J}^\top \mathbf{J} \mathrm{d}\Omega\right) \dot{\mathbf{q}} \\
\approx \frac{1}{2} \dot{\mathbf{q}}^\top \left[ \sum_{\mathbf{x}_k \in \mathcal{I}} \rho V_k \mathbf{J}^\top(\mathbf{x}_k)\mathbf{J}(\mathbf{x}_k) \right]\dot{\mathbf{q}},
\end{multline}
and $\mathbf{M} = \sum_{\mathbf{x}_k \in \mathcal{I}} \rho V_k \mathbf{J}^\top(\mathbf{x}_k)\mathbf{J}(\mathbf{x}_k) $ is the mass matrix. Note that $V_k = h_1 h_2 h_3$ is the estimated volume of the IP cuboid, not the volume of the PDS particle. $\rho$ is the density of the 3D model, which should not be confused with $\sigma$.

\begin{figure*}
  \centering 
  \includegraphics[width = \linewidth]{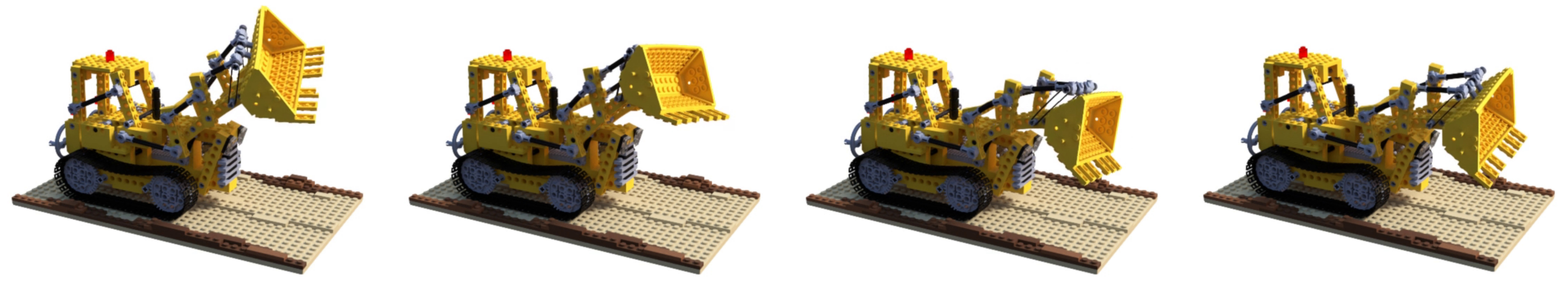}
  \caption{\textbf{Elastically deforming excavator.}~The excavator is a standard benchmark for NeRF-based frameworks. We use this classic model to showcase the capability of PIE-NeRF, which generates interesting and novel dynamic effects in real time.}
  \label{fig:exc}
\end{figure*}

Integrating the potential energy $U$ is handled in a similar way. Under the assumption of hyperelasticity, the energy density $\Psi(\mathbf{x})$ depends on the deformation gradient at $\mathbf{x}$: $\mathbf{F} = \nabla \mathbf{\mathbf{u}(\mathbf{x})+ \mathbf{I}}\in\mathbb{R}^{3 \times 3}$. According to Eq.~\eqref{eq:u}, the deformation gradient at the $k$-th IP is:
\begin{equation}\label{eq:F}
    \mathbf{F}(\mathbf{x}_k) = \mathbf{F}_k  = \mathbf{q} \cdot \nabla \mathbf{J}^\top(\mathbf{x}_k)  + \mathbf{I} = \mathbf{q} \cdot \nabla \mathbf{J}^\top_k + \mathbf{I}.
\end{equation}
$\mathbf{J}_k \in \mathbb{R}^{3 \times 30n}$ is the Jacobi corresponding to the IP, and $\nabla \mathbf{J}^\top_k \in \mathbb{R}^{30n \times 3 \times 3}$ is a third tensor. The potential accumulated at the IP is estimated by integrating over its cuboid ($\Omega_k$) assuming the IP lies at the center:
\begin{equation}
    U_k = \int_{\Omega_k}\Psi(\mathbf{F}(\mathbf{h})) = \int_{-\frac{h_1}{2}}^{\frac{h_1}{2}} \int_{-\frac{h_2}{2}}^{\frac{h_2}{2}} \int_{-\frac{h_3}{2}}^{\frac{h_3}{2}} \Psi(\mathbf{F}(\mathbf{h})).
\end{equation}
Here, $\mathbf{h}$ is the local coordinate spanning $\Omega_k$, and $\mathbf{x}_k$ aligns with $\mathbf{h} = 0$. When $\mathbf{h}\neq 0$, we first-order approximate $\mathbf{F}$ as:
\begin{equation}\label{eq:appro_F}
\mathbf{F}(\mathbf{h})\approx \mathbf{F}(0) + \nabla \mathbf{F}(0) \cdot \mathbf{h} = \mathbf{F}(\mathbf{x}_k) + \nabla \mathbf{F}(\mathbf{x}_k) \cdot \mathbf{h}.
\end{equation}
This makes sense because Q-GMLS assumes $\mathbf{u}$ is quadratic, which has a linearly-vary deformation gradient. Therefore, the approximate in Eq.~\eqref{eq:appro_F} should be exact. $\nabla \mathbf{F}$ can be computed by differentiating Eq.~\eqref{eq:F}:
\begin{equation}\label{eq:nablaF}
    \nabla \mathbf{F}(\mathbf{x}_k) =  \mathbf{q} \cdot \nabla^2 \mathbf{J}^\top_k  = \mathbf{H}_k \cdot \mathbf{q}.
\end{equation}
Here $\mathbf{H} = \partial \nabla \mathbf{J}^\top/\partial \mathbf{x}$ is a fourth tensor. This computation boils down to evaluating first- and second-derivatives of $N_i$, $N_i^j$, and $N_i^{jk}$, and can be pre-computed per Eq.~\eqref{eq:N}. 

Given an elastic material model $\Psi(\mathbf{F})$, the total potential can then be computed via:
\begin{equation}
    U \approx \sum_{\mathbf{x}_k \in \mathcal{I}}\int_{\Omega_k} \Psi\left(\mathbf{F}(0) + \mathbf{H}_k : (\mathbf{q} \mathbf{h}^\top)\right) \mathrm{d}\Omega_k.
\end{equation}
The actual integration computation relies on the specific formulation of $\Psi(\mathbf{F})$. Please refer to the supplementary document for detailed derivations of some commonly-used material models such as ARAP and Neo-Hookean.

\subsection{System assembly and solve}
With energy integrals, we can assemble Eq.~\eqref{eq:motion}. The generalized internal force is $\mathbf{f}_{int} = -\partial U / \partial \mathbf{q}$, and it can be conveniently computed by the chain rule:
\begin{equation}\label{eq:f_int}
\mathbf{f}_{int} = - \frac{\partial \Psi}{ \partial \mathbf{F}} : \frac{\partial \mathbf{F}}{ \partial \mathbf{q}} = - \mathbf{P} : \nabla \mathbf{J},
\end{equation}
where $\mathbf{P}$ is the first Piola-Kirchhoff stress. This is a $30n$-dimension dense system as all Q-GMLS kernels have global influences. We use Newton's method to solve this system iteratively. Each Newton iteration solves a linearized problem for an incremental improvement $\Delta \mathbf{q}$:
\begin{equation}\label{eq:newton}
    \left(\mathbf{M} + h^2 \frac{\partial \mathbf{f}_{int}}{\partial \mathbf{q}} \right)\Delta\mathbf{q} = \mathbf{M}(\mathbf{q}_{n} + h\dot{\mathbf{q}}_n) + h^2\mathbf{J}^\top\mathbf{f}_{ext},
\end{equation}
where $\partial \mathbf{f}_{int} / \partial \mathbf{q}$ is the second differentiation of the total potential $U$ known as the tangent stiffness matrix. $\mathbf{f}_{ext}$ is the external forces applied to the model. It is projected to the Q-GMLS kinematic space by left multiplying $\mathbf{J}^\top$. 

\subsection{NeRF rendering using quadratic warping}\label{subsec:warping}
After the deformed model geometry is computed, we leverage the NGP-NeRF that is built for its rest shape to synthesize \emph{both novel views and novel deformations}. Whenever we query the NGP-NeRF for a deformed location $\tilde{\mathbf{x}}$ along a ray, we warp this position to its rest configuration $\mathbf{x}$, ideally through $\mathbf{x} = \tilde{\mathbf{x}} - \mathbf{u}(\mathbf{x})$. Unfortunately as $\mathbf{x}$ is unknown here, we cannot obtain $\mathbf{u}(\mathbf{x})$ directly with Eq.~\eqref{eq:u}. Instead, we approximate $\mathbf{u}(\mathbf{x})$ based on the displacements at nearby (deformed) IPs. The general rationale is that if $\tilde{\mathbf{x}}$ is sufficiently close to an IP, we can Taylor expand the IP's displacement to estimate $\mathbf{u}(\mathbf{x})$. As IPs are sparse, it is possible that $\tilde{\mathbf{x}}$ is not particularly close to one IP. In this case, we find three nearest IPs and average Taylor expansions at those IPs based on the inverse distance weight.

For the IP at $\mathbf{x}_k$, we have:
\begin{multline}\label{eq:warp}
    \tilde{\mathbf{x}} - \tilde{\mathbf{x}}_k = \mathbf{u}(\mathbf{x}) - \mathbf{u}(\mathbf{x}_k) \approx \nabla \mathbf{u}(\mathbf{x}_k) (\mathbf{x} - \mathbf{x}_k) \\
    + \frac{1}{2} \left(\nabla\mathbf{F}(\mathbf{x}_k) \cdot (\mathbf{x} - \mathbf{x}_k)\right)\cdot(\mathbf{x} - \mathbf{x}_k).
\end{multline}
We can then compute $\mathbf{x}$ via solving a nonlinear system of:
\begin{equation}\label{eq:warpsys}
    \mathbf{A} (\mathbf{x}) (\mathbf{x}-\mathbf{x}_k) = \mathbf{b}, 
\end{equation}
where 
\begin{equation}
\begin{array}{l}
\displaystyle \mathbf{A}(\mathbf{x}) = \nabla \mathbf{u}(\mathbf{x}_k) + \frac{1}{2} \nabla \mathbf{F}(\mathbf{x}_k) \cdot (\mathbf{x} - \mathbf{x}_k),\\ 
\displaystyle \mathbf{b} = \tilde{\mathbf{x}} - \tilde{\mathbf{x}}_k.
\end{array}
\end{equation}
While the analytic solution of Eq.~\eqref{eq:warpsys} can be derived, we find Newton's method starting from the guess of $\mathbf{x} = \mathbf{x}_k$ is effective. The system converges within tens of iterations, and each iteration only solves a 3 by 3 linear system. 

This strategy of quadratic warping fully exploits the prior of $\mathbf{u}$ being the quadratic displacement field. If we only use the first-order Taylor expansion to estimate the undeformed position of $\tilde{\mathbf{x}}$, as chosen in most existing NeRF editing systems~\cite{peng2022cagenerf,xu2022deforming}, visual artifacts can be observed under large deformations. Examples of such failure cases are provided in the supplementary materials.

\section{Experiments}\label{sec:exp}
We implemented PIE-NeRF pipeline using \texttt{Python} and \texttt{C++}. The simulation module was based on \texttt{CUDA}. In addition, we used \texttt{PyTorch}~\cite{imambi2021pytorch} and \texttt{Taichi}~\cite{hu2019taichi} to implement a modified instant-NGP~\cite{Muller2022ngp} for ray warping (\S~\ref{subsec:warping}). Our hardware platform is a desktop computer equipped with an \texttt{Intel} \texttt{i7-12700F} CPU and an \texttt{NVIDIA} \texttt{3090} GPU.

\noindent\textbf{Datasets} 
We evaluate PIE-NeRF with several NeRF scenes. In addition to original NeRF datasets, we utilize BlenderNeRF~\cite{raafat2023blendernerf} to synthesize additional scenes including codimensional objects. We used 100 multi-view images for each scene as inputs for our NGP-NeRF training. 

% In this section, we will first show that our model can generate high quality dynamics on different scenes. As a proof of unified simulator, we also demonstrate our model's codimensional simulation both on rod and shell. Last, we will demonstrate the cost of every component in our model which help to achieve interactive rate and enable user to generate desiring dynamics.
\begin{figure}
  \centering 
  \includegraphics[width= \linewidth]{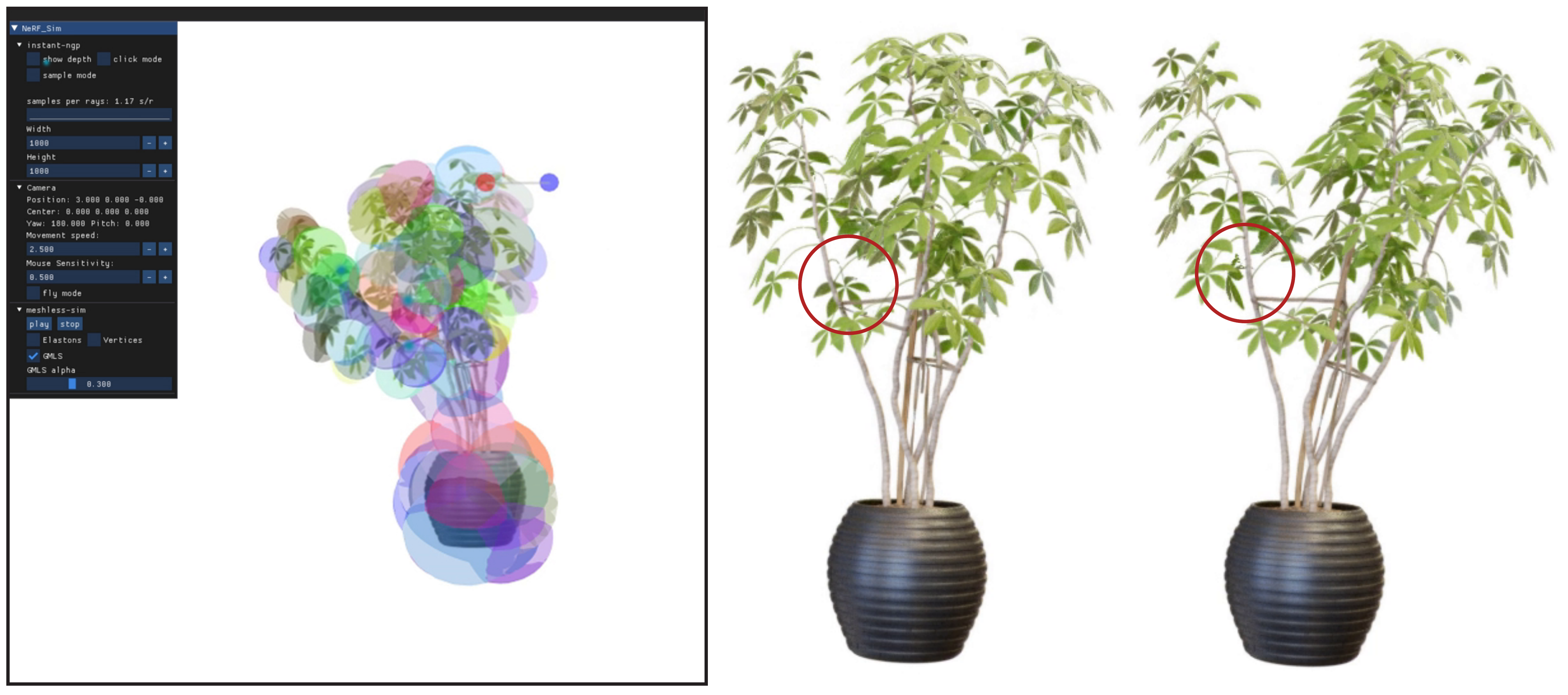}
  \caption{\textbf{Interactive NeRF deformation.}~We developed an intuitive UI for users to interact with NeRF scenes like applying external forces and position constraints. Q-GMLS kernels can also be set adaptively to capture local dynamics as highlighted.}
  \label{fig:ui}
  \vspace{-3 pt}
\end{figure}

\subsection{Interactive and dynamic NeRF simulation}
PIE-NeRF formulates nonlinear dynamics of NeRF models with the generalized coordinate and Lagrangian equations (i.e., Eq.~\eqref{eq:lagrangian}), which makes the computation independent of the PDS sampling resolution. We find that a few dozen Q-GMLS kernels are often sufficient to model complex models. The corresponding computation is light-weight and can be processed in real-time on the GPU (see Fig.~\ref{fig:exc}). Therefore, interactive physics-based manipulation of the NeRF scene becomes possible. To this end, we also implemented a user-friendly interface as shown in Fig.~\ref{fig:ui}, left. With the interface, users can intuitively apply external forces to the model, and observe the resulting novel motions interactively. The users have full control over the trade-off between visual richness and the efficiency of the simulation. For instance, one can create a dedicated kernel to capture local dynamics at specific foliage of the plant (Fig.~\ref{fig:ui}, right). 

\begin{figure}
  \centering 
  \includegraphics[width=0.95\linewidth]{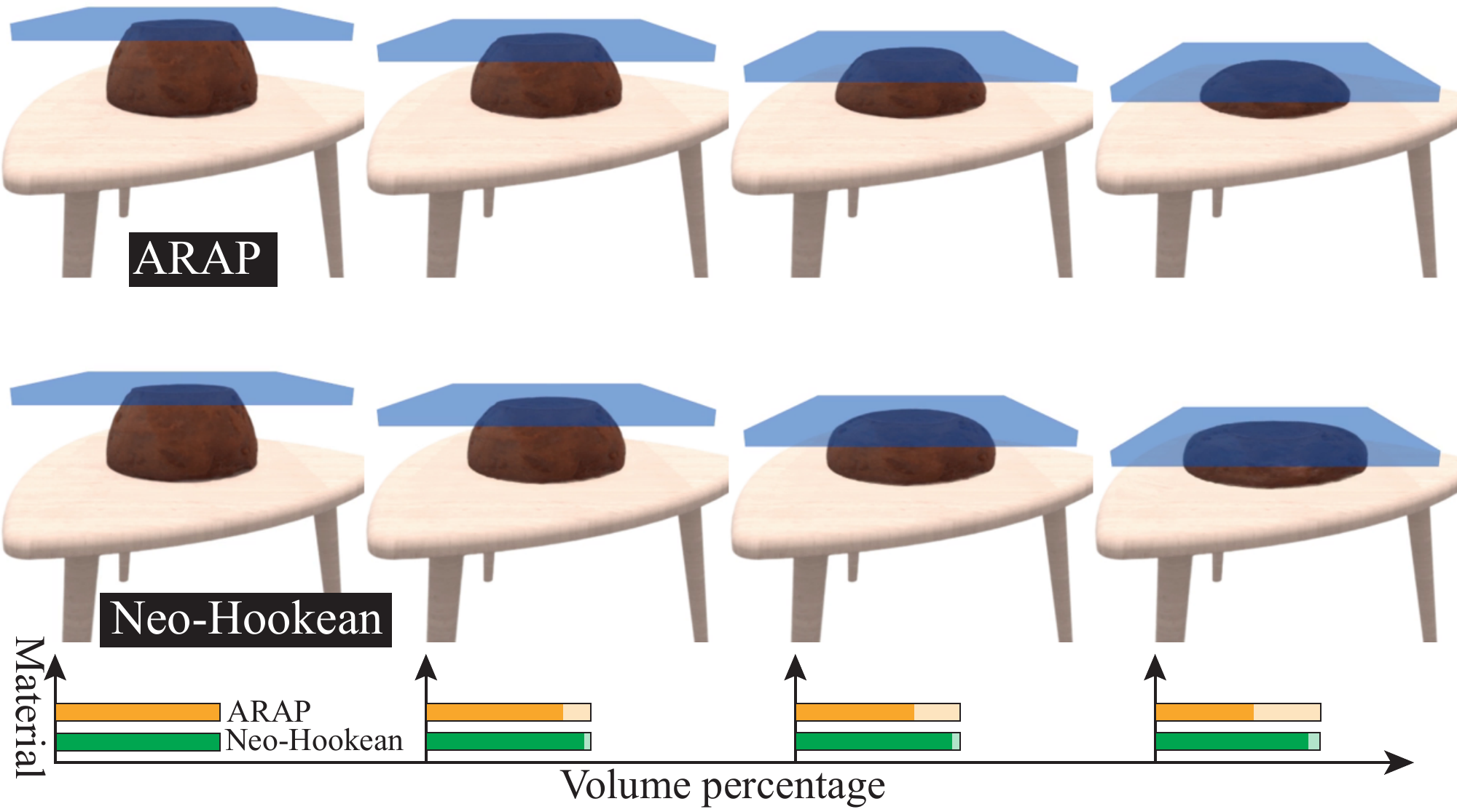}
  \caption{\textbf{Volume preservation test.}~PIE-NeRF is capable of incorporating any real-world material models. In this example, we apply a position constraint to compress a chocolate jelly. ARAP energy, widely used in exiting NeRF editing systems~\cite{yuan2022nerf}, collapses unnaturally. With Neo-Hookean energy, PIE-NeRF can better synthesize this procedure. }
  \label{fig:compress}
  \vspace{-5 pt}
\end{figure}
\subsection{Physics-grounded pose synthesis}
Being a physics-based framework, PIE-NeRF is able to model any nonlinear hyperelastic materials to match real-world observations. This enhances existing NeRF editing systems, which are mostly based on geometry-based heuristic energy models like ARAP. Fig.~\ref{fig:compress} reports a comparison using the Neo-Hookean material~\cite{bonet1997nonlinear} and ARAP to compress chocolate jelly with NeRF. The energy density of Neo-Hookean material is:
\begin{equation}\label{eq:nh}
\Psi=\frac{\mu}{2}\left(I_C-3\right)-\mu \log J+\frac{\lambda}{2} \log ^2J, 
\end{equation}
where $\mu$, $\lambda$ are material parameters (a.k.a Lam\'{e} coefficients); $I_C = \text{tr}(\mathbf{F}^\top\mathbf{F})$; and $J = \text{det}(\mathbf{F})$. The log barrier $\log J$ in Neo-Hookean energy strongly preserves the volume of the object. This feature is clearly demonstrated in Fig.~\ref{fig:compress}. As the compression rate increases, ARAP jelly (on the top) loses nearly $40\%$ of the original volume (visualized as bar graphs at the bottom in Fig.~\ref{fig:compress}).

\begin{figure}
  \centering 
  \includegraphics[width=0.95\linewidth]{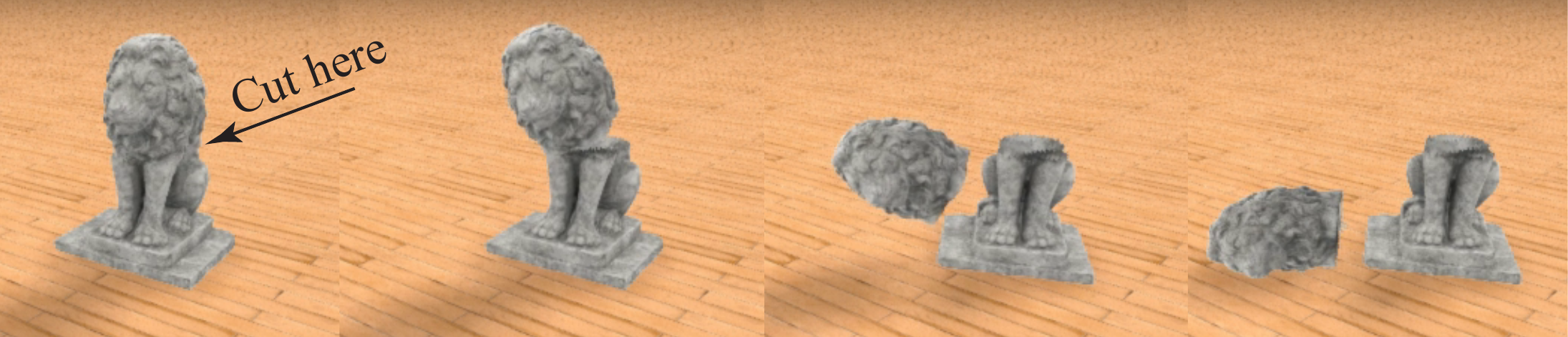}
  \caption{\textbf{Topology change and shadows.}~We edit kernel weights to cut the sculpture, which then falls on the floor. We then compute the depth map from NGP-NeRF and generate the moving shadow using shadow maps. }
  \label{fig:cut}
  \vspace{-10 pt}
\end{figure}
\vspace{3 pt}
\noindent\textbf{Handle topology change.}~
Being meshless makes PIE-NeRF less sensitive to topology changes. As shown in Fig.~\ref{fig:cut}, we cut the NeRF sculpture by modifying Q-GMLS weight functions. Besides that, there is no need for extra safeguards for dealing with the change of the mesh connectivity and resolution at the cutting area. In this example, we extract the depth image with NGP-NeRF, from which a shadow map can be generated for shadow synthesis. 

\vspace{3 pt}
\noindent\textbf{Comparison with ground truth.}
We employ FEM as the ground truth for comparison with our method. We generate multi-view images from the mesh rendered with same rendering settings to ground truth as the dataset for PIE-NeRF. Subsequently, dynamic results are produced using the same boundary conditions, physical parameters, and external forces. Our implementation of FEM involves tetrahedralizing the mesh and applying Newton's method to solve the dynamic system on tetrahedral mesh. The comparison results, as illustrated in Fig.~\ref{fig:gtcomp}, demonstrate that our method is nearly same to the ground truth. Thanks to the Q-GMLS, the number of kernels in our approach is significantly lower than the number of vertices in FEM, and the number of IPs is also less than the number of tetrahedra in FEM.

% We compare our method with the ground truth following the procedures below. Given a textured surface mesh, we generate an underlying tetrahedral mesh for FEM simulation. Using the Newton's method, we solve for dynamics of this tetrahedral mesh and interpolate to obtain deformed surface mesh. The deformed surface meshes are further rendered with Blender to get the results. As for our method, we generate NeRF dataset with this textured surface mesh, the same as any synthetic data using Blender, and then follow our pipeline described in the paper. We make sure two things. First, the resolution of the tetrahedra used in FEM simulation and the density of IPs used in our method are similar. Second, the boundary conditions are the same, including external forces and fixed points. The comparisons are shown in Fig.~\ref{fig:gtcomp}. We can see that our dynamics is similar to the ground truth.

\begin{figure}
  \centering 
  \includegraphics[width =\linewidth]{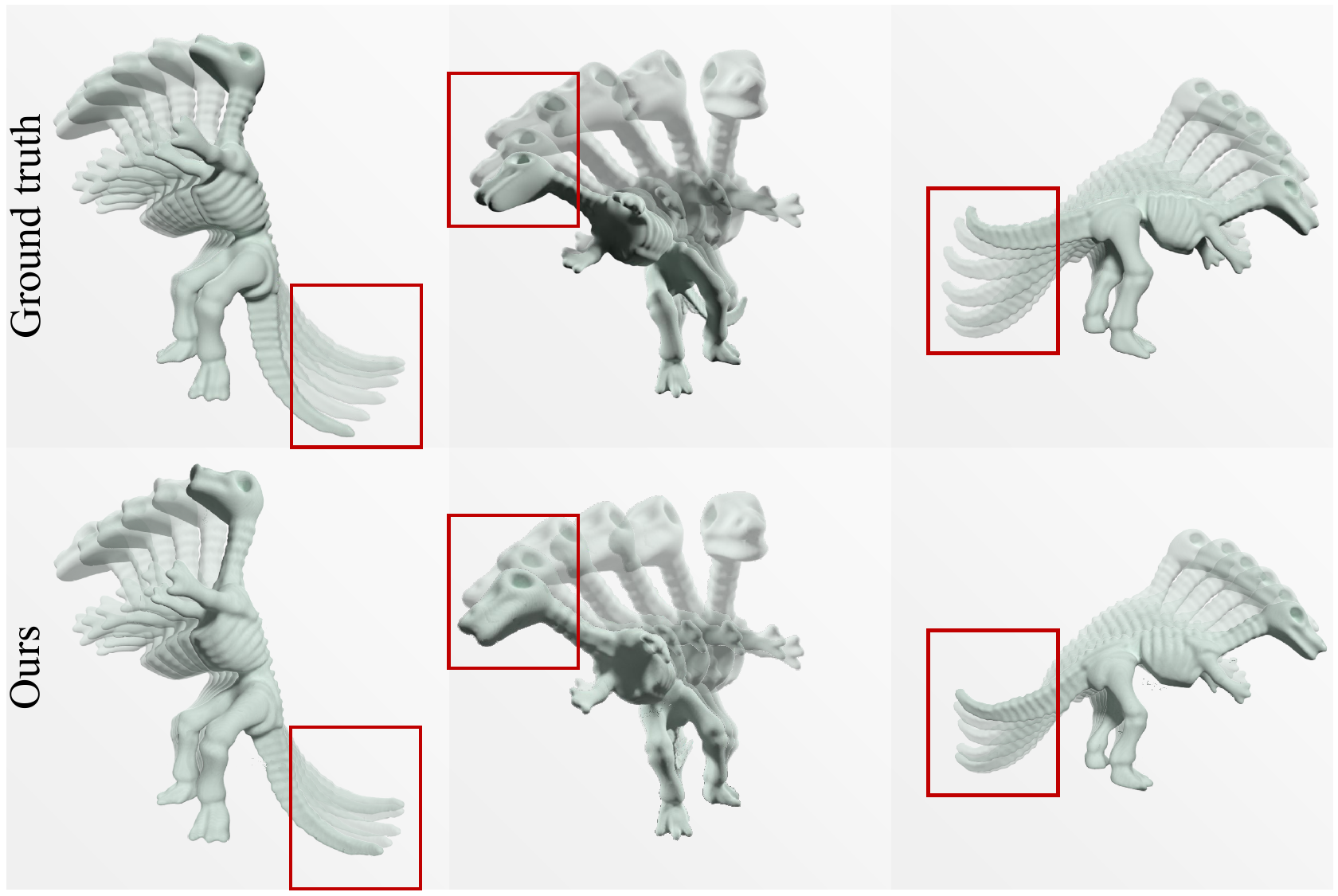}
  \caption{\textbf{Comparison with ground truth.}~We generate ground-truth results by simulating the tetrahedral mesh with FEM. We can see that our results are quite similar to the ground truth, despite minor differences highlighted.}
  \label{fig:gtcomp}
  \vspace{-7pt}
\end{figure}

\vspace{3 pt}
\noindent\textbf{Comparison with PAC-NeRF.} 
PAC-NeRF~\cite{li2023pac} is a recent contribution also aiming to combine physical models with NeRF-based representations. The underlying numerical solver, on the other hand, is based on the material point method (MPM)~\cite{bardenhagen2004generalized}, a hybrid method that uses both particles and grids. While MPM excels in handling complicated, multi-phase physics, it does not synergize well with NeRF-based rendering. Specifically, under large deformation, PAC-NeRF fails to map material points back to their rest-pose positions accurately due to excessive interpolation smoothing between particles and grid cells, which leads to over-blurred results at local fine shapes. We show a side-by-side comparison in Fig.~\ref{fig:PAC}.

\subsection{Time performance}

\begin{figure}
  \centering 
  \includegraphics[width=0.93\linewidth]{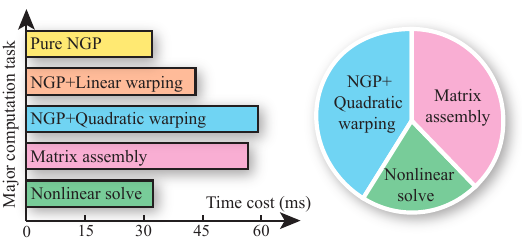}
  \caption{\textbf{Time breakdown.}~There are three major steps for PIE-NeRF, namely matrix assembly, solve, and warping. Thanks to Q-GMLS-based reduction, they are all manageable making the simulation interactive. The percentage of each step is visualized as a pie diagram on the right.}
  \label{fig:timecost}
  \vspace{-7pt}
\end{figure}

Fig.~\ref{fig:timecost} reports a breakdown of the run time performance of our PIE-NeRF pipeline for Fig.~\ref{fig:ui}. Three major tasks for the simulation at the runtime are matrix assembly, nonlinear solve, and quadratic warping. The matrix assembly requires an integral over all the IPs, leading to a dense $30n$ by $30n$ system. We use Cholesky factorization to solve the resulting Newton system (Eq.~\eqref{eq:newton}). In general, a couple of iterations will converge the system so that we forward to the next time step. As shown in the figure, quadratic warping used in PIE-NeRF is slightly more expensive than linear warping. However, these additional expenses yield significantly improved visual results in general, as detailed in the supplementary material.

\begin{figure}
  \centering 
  \includegraphics[width=0.99\linewidth]{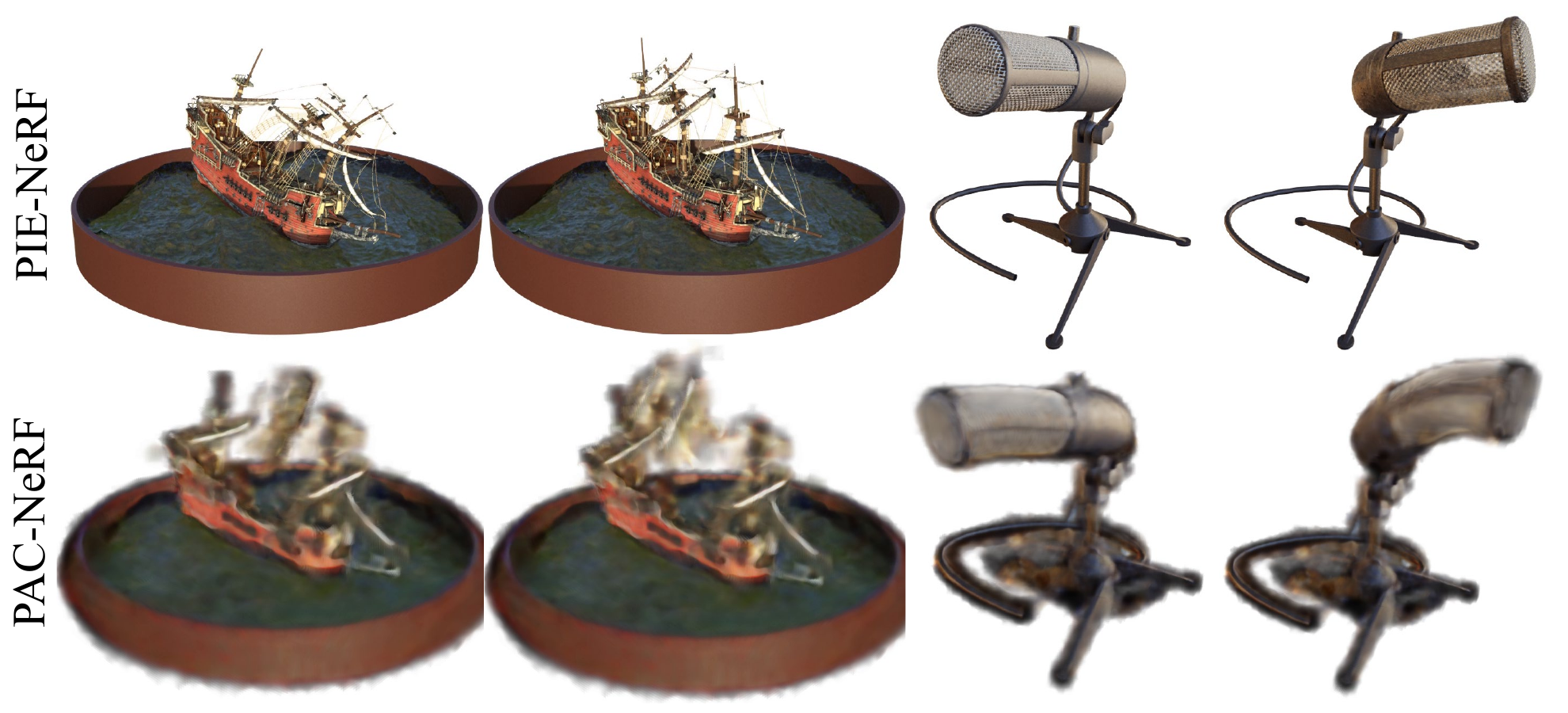}
  \caption{\textbf{PIE-NeRF vs PAC-NeRF.}~PAC-NeRF~\cite{li2023pac} is a closely relevant competitor. A major limitation of PAC-NeRF is the rendering. It is less intuitive to infer the right color/texture information under large deformation using MPM. PIE-NeRF overcomes this limitation with ease. Using implicit time integration, PIE-NeRF runs much faster than PAC-NeRF (two orders in these examples).}
  \label{fig:PAC}
  \vspace{-10 pt}
\end{figure}

% \subsection{Evaluation of interactive rate}
% \noindent\textbf{Interactive user interface}
% To enable user generate desire dynamics, we designed an user interface that can visualize GMLS kernels and drag particlular parts to add external force which deform the object to specific direction.
% \noindent\textbf{Performance}
% Our performance is tested on Ficus scene in Synthesis-NeRF datasets. \ref{fig:interactiveperformance} shows the average timimngs for per-frame matrix assembly, per-frame solution of the nonlinear system ,per-frame NGP rendering and both linear and quadratic ray warping. As a result, our efficient model can achieve interactive rate to generate and interact with an elastodynamics.

\section{Conclusion}\label{sec:conclusion}
PIE-NeRF\faPizzaSlice~is a physics-NeRF simulation pipeline. It is directly based on PDS particles sampled over the NeRF scene and applies a Q-GMLS model reduction to lower the computational overhead of the simulation. As a result, PIE-NeRF faithfully characterizes various real-world material models. Its meshless representation makes the simulation flexible, and topology changes can be well accommodated. The quadratic interpolation scheme is not only helpful in tackling thin-geometry models but also leads to better image synthesis with NGP-NeRF. We hope PIE-NeRF could contribute new ingredients to the existing NeRF ecosystem. Based on PIE-NerF, it is possible to integrate more (better and faster) simulation and graphics techniques to deep 3D vision applications to imbue vivid, realistic, real-time physics into static or dynamic environments. Along this exciting endeavor, we will also explore other opportunities, such as Gaussian splatting-based techniques~\cite{kerbl20233d}, and ultimately reach \emph{what you see is what you simulate,  WS\textsuperscript{2}}.
\section*{Acknowledgement}\label{sec:acknowledgement}
We thank anonymous reviewers for their insightful comments. We acknowledge support from NSF (2301040, 2008915, 2244651, 2008564, 2153851, 2023780), NSF-China (62322209), UC-MRPI, Sony, Amazon, and TRI. 
%------------------------------------------------------------------------
%\section{Final copy}

% You must include your signed IEEE copyright release form when you submit your finished paper.
% We MUST have this form before your paper can be published in the proceedings.

%%%%%%%%% REFERENCES
{\small
\bibliographystyle{ieee_fullname}
\bibliography{egbib}
}

% \appendix
% \input{sup}
%\input{sup}

\end{document}